\documentclass[10pt,journal,compsoc]{IEEEtran}

\usepackage{flushend}
\usepackage{xcolor}
\usepackage{amsmath}
\usepackage{amsfonts}
\usepackage{tabularx}
\usepackage{graphicx}

\begin{document}

\title{Optimizing for In-memory Deep Learning with Emerging Memory Technology}


\author{Zhehui~Wang, Tao~Luo, Rick~Siow~Mong~Goh, Wei~Zhang, Weng-Fai~Wong
\thanks{Zhehui Wang, Tao Luo and Rick~Siow~Mong~Goh are with the Institute of High Performance Computing (IHPC), Agency for Science, Technology and Research (A*STAR), Singapore.}
\thanks{Wei Zhang is with the Hong Kong University of Science and Technology.}
\thanks{Weng-Fai Wong is with the National University of Singapore.}
}


\IEEEtitleabstractindextext{

\begin{abstract}

In-memory deep learning computes neural network models where they are stored, thus avoiding long distance communication between memory and computation units, resulting in considerable savings in energy and time. In-memory deep learning has already demonstrated orders of magnitude higher performance density and energy efficiency. The use of emerging memory technology promises to increase the gains in density, energy, and performance even further. However, emerging memory technology is intrinsically unstable, resulting in random fluctuations of data reads. This can translate to non-negligible accuracy loss, potentially nullifying the gains. In this paper, we propose three optimization techniques that can mathematically overcome the instability problem of emerging memory technology. They can improve the accuracy of the in-memory deep learning model while maximizing its energy efficiency. Experiments show that our solution can fully recover most models' state-of-the-art accuracy, and achieves at least an order of magnitude higher energy efficiency than the state-of-the-art.


\end{abstract}

}

\maketitle

 

\IEEEpeerreviewmaketitle

\section{Introduction}

Deep learning neural networks (DNNs) are widely used today in many applications such as image classification, object detection, etc. High throughput and high energy efficiency are two of the pressing demands on DNNs these days. However, current computation mechanisms are not the best choices for DNN in terms of efficiency. As DNN models become increasingly complex, the computation process takes a lot of time and energy~\cite{inference2015performance}. In traditional von Neumann architectures, storage memory is separated from computation operations. To compute a output feature map, we need to read the input feature map and weights from the memory units, send them to the computation module to compute before writing the results back to the memory units. During the whole process, the system spends a large portion of energy and time in data movement~\cite{yang2017designing}. This is made worse with advances in process technology, making the relative distances involved even longer and thus more costly.

{\em Emerging memory technologies} (EMT) including PCRAM, STT-RRAM and FeRAM~\cite{meena2014} promise better density and energy efficiency, especially since many are non-volatile and well suited to the read-mostly applications. 
In-memory deep learning using EMT cells, especially in {\em analog} mode, has already demonstrated an order of magnitude better performance density and two orders of magnitude better energy efficiency than the traditional deep learning on the MNIST dataset~\cite{yao2020fully}~\cite{cai2019fully}~\cite{luo2021nc}.
As in-memory deep learning integrates computation with the memory operations~\cite{sebastian2020memory}, the computation results can be directly read from the memory modules using a single instruction.  This is different from traditional deep learning, where the memory operations are executed separately from the computation operations. Computing where data is stored reduces the need to move large amounts of data around frequently. Especially when technology scales and on-chip distances become longer, we can expect substantial savings of time and energy if the emerging memory technology is used in the in-memory deep learning paradigm. 

A big challenge of in-memory computing is the instability of EMT cells~\cite{raghavan2013rtn}~\cite{luo2018fpga}. Unlike traditional memory technology, where the data are stored in stable memory cells, data stored in analog mode EMT cells may fluctuate and different values can be output. For instance, suppose we store the weight $w$ in an EMT cell. When we read it from the EMT cell, the output may become $w+\Delta w$ instead of $w$. Here $\Delta w$ is a fluctuating amplitude of that memory cell. Because of this instability, in-memory deep learning using EMT especially in an analog manner may make incorrect classification~\cite{du2020exploring}. This can severely limit its application in the real world.

Another challenge of in-memory deep learning is the ineffectiveness of traditional energy reduction techniques, such as pruning, quantization, etc. Pruning reduces energy consumption by decreasing the number of operations~\cite{yang2017designing}~\cite{DeepCompression}. However, the energy of in-memory deep learning is roughly proportional to the weight values. Pruning usually removes those weights with smaller absolute values, which only contribute to a small portion of the energy consumption. Quantization reduces the energy consumption by decreasing the complexity of operations using low precision data~\cite{wang2019haq}~\cite{wang2021evolutionary}. Unfortunately, in EMT cells, low precision data consumes almost the same amount of energy as that of the high-precision data. Therefore, traditional pruning and quantization technologies are less effective for in-memory deep learning.

In this paper, we solve these problems by proposing three techniques. They can effectively recover the model accuracy loss in in-memory deep learning and minimize energy consumption. Our innovations are:
\begin{itemize}
\item\textbf{Device enhanced Dataset:} We propose augmenting the standard training dataset with device information integrated into the dataset. In this way, the models will also learn the fluctuation patterns of the memory cells. This extra information can also help the optimizer avoid overfitting during training and thus improve the model's accuracy.
\item \textbf{Energy Regularization:} We optimize the energy coefficients of memory cells and the model parameters by adding a new term into the loss function. This new term represents the energy consumption of the model. The optimizers can automatically search for the optimal coefficients and parameters, improving the model accuracy and energy efficiency at the same time. 
\item \textbf{Low fluctuation Decomposition:} We decompose the computation into several time steps. This is customized for in-memory deep learning and can compensate for the fluctuation of memory cells. This decomposed computation mechanism can improve the model accuracy and energy efficiency.
\end{itemize}
We prove the effectiveness of all these techniques both theoretically and experimentally. They can substantially improve the model accuracy and energy efficiency of models for in-memory deep learning. We organize the rest of the paper as follows: Section~\ref{s2} gives related work; Section~\ref{s3} presents the background knowledge; Section~\ref{s4} introduces our proposed optimization method; Section~\ref{s5} shows the experiment results; Section~\ref{s6} concludes.

\begin{figure*}[!t]
  \centering
  \includegraphics[width=7.0in] {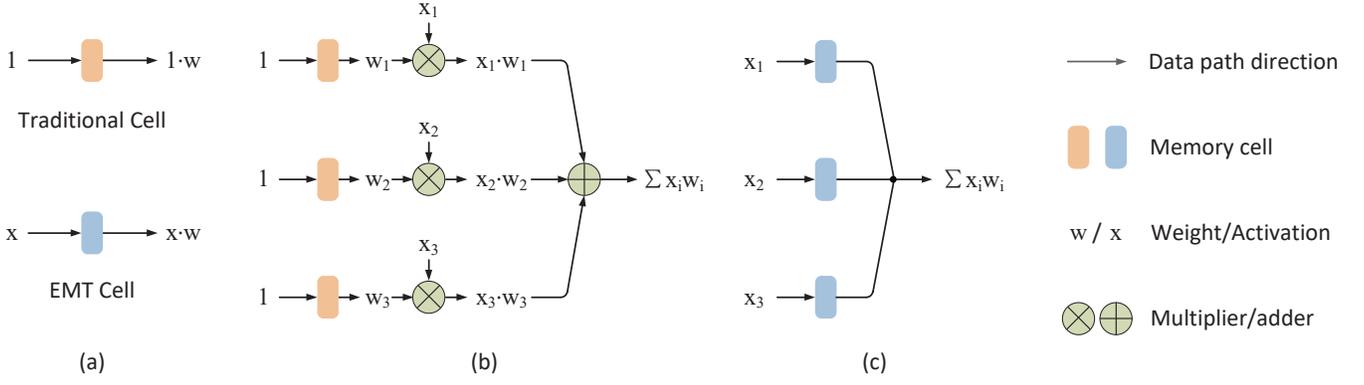}
  \caption{(a) Comparison between the traditional memory cell and the EMT cell; (b) The computation mechanism for traditional deep learning; (c) The computation mechanism for EMT-based in-memory deep learning, where we integrate multiplication and addition into the read operations.}
  \label{f:work_overview}
\end{figure*}

\begin{figure}[!t]
  \centering
  \includegraphics[width=3.5in] {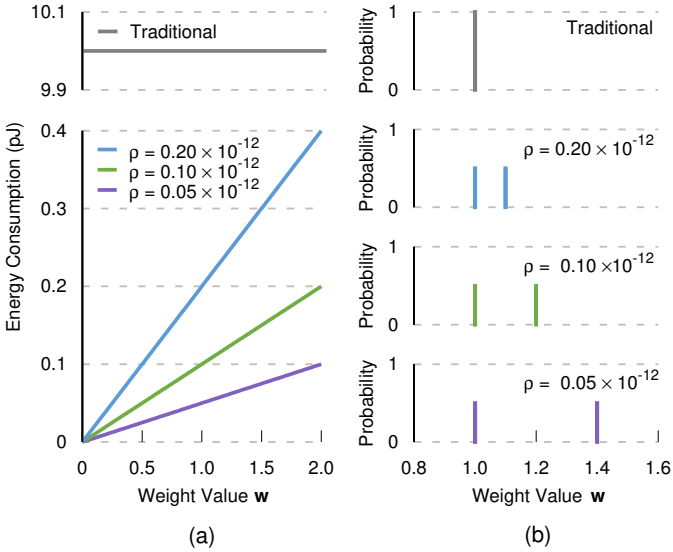}
  \caption{(a) The relationship between the energy consumption of the EMT cell and the value of the weight; (b) The probability distribution of the weight fluctuation, with different energy coefficients $\rho$. The traditional cell is marked in grey color for reference.}
  \label{f:basic}
\end{figure}

\section{Related Work}
\label{s2}

The first category of work to improve the accuracy of the in-memory deep learning model is called \textit{binarized encoding}. The information in each memory cell is digitized into one bit instead of being stored as a full precision number. In other words, the data stored in each memory cell is either $1$ or $0$~\cite{emara2014differential}. Theoretically, the one-bit data is more robust than a high-precision number at the same level of fluctuation.  Several previous works used the one-bit design to compute the binarized neural networks. Sun \textit{et al.} used the single-bit cell to execute the XNOR operations in XNOR-Net~\cite{sun2018xnor}. Chen \textit{et al.}~\cite{chen201865nm} used the single-bit cell to execute the basic operations in binary DNN. Tang \textit{et al.} proposed a customized binary network for the single-bit cell ~\cite{tang2017binary}. However, either the XNOR-Net or the binary neural network has a large accuracy drop compared with the full-precision model~\cite{rastegari2016xnor}. Recently, new progress in this research direction is to store a high-precision weight using a group of single-bit memory cells. For example, Zhu \textit{et al.}~\cite{zhu2019configurable} used single-bit memory cells $\times N$ to store a $N$-bit weight. Such a method can increase the model accuracy because it can increase the effective precision of weights. Compared with the traditional design, it uses more memory cells.

The second category of work to improve the accuracy of the in-memory deep learning model is called \textit{weight scaling}. Theoretically, we can reduce the amplitude of weight fluctuation by scaling up the weight values stored in the memory cell~\cite{peng2019optimizing}. After computation, we scale the result down using the same scaling factor. In the literature, many research works have found other physical ways to reduce the amplitude of weight fluctuation. For example, He \textit{et al.} found that we can reduce the fluctuation amplitude by lowering the operation frequency~\cite{he2019noise}. Chai \textit{et al.} found one material, which has lower fluctuation amplitude than the other types of material~\cite{chai2018impact}. However, these methods demand strict physical conditions. Compared with them, weight scaling is a more general method that can reduce the fluctuation amplitude of memory cells in most conditions~\cite{sorbaro2020optimizing}. However, Choi \textit{et al.} found that although the memory cell using scaled weights showed smaller fluctuation amplitude, it also consumed higher energy consumption~\cite{choi2014random}. Ielmini \textit{et al.} modeled the relationship between the scaling factor, the fluctuation amplitude, and the energy consumption~\cite{ielmini2010resistance}, which could help us to find the optimal scaling factor for in-memory computing computation.

The third category of work to improve the accuracy of the in-memory deep learning model is called \textit{fluctuation compensation}. To alleviate the instability, they first read the memory cell by multiple times and then record the statistical result such as mean and standard deviation~\cite{puglisi2015statistical}. Afterward, they either calibrate the model parameter or the model output directly based on that statistical results~\cite {shim2020two}. This method is also widely used during the memory cell programming stage. For example, Joshi \textit{et al.} compensated the programming fluctuation by tuning the batch normalization layer parameters~\cite{joshi2020accurate}.
Alternatively, Zhang \textit{et al.} compensated the programming fluctuation by offsetting the weight values~\cite{zhang2020reliable}. These methods are effective in a static device environment. If we face a dynamic environment, a more general way is needed. One popular approach is to have many equivalent models running in parallel. Then we calculate the mean of the results. Joksas \textit{et al.} did this by applying the committee machine theory into the in-memory computing devices~\cite{joksas2020committee}. Wan \textit{et al.} optimized this process by running a single model on the same device and reading the memory cells multiple times~\cite{wan2020voltage}. This method can average out the weight fluctuation and get a more stable result.

\section{Preliminaries}
\label{s3}

EMT-based in-memory deep learning can be very efficient~\cite{yao2020fully} because of its analog operation. Fig.~\ref{f:work_overview}(a) shows the difference between the traditional and EMT memory cells. When we read a weight $w$ from the traditional memory cell, the input to the corresponding memory cell is $1$, meaning that the read request to that memory cell is enabled. Afterward, the memory cell returns $w$ as an output. An EMT cell for in-memory deep learning is quite different. When we read the weight $w$, the input to the memory cell is a variable $x$ instead of the fixed data $1$. The memory cell then returns $x\cdot w$ directly, the product of the input signals $x$ and the stored weight $w$. In other words, the EMT cell integrates the multiplication operation into the read operation.

Analog EMT is more efficient than traditional memory not only in multiplication operations but also in addition operations~\cite{pedretti2021memory}. To better explain this, we show how traditional cells and EMT cells execute the {\em multiply-accumulate} (MAC) operation in Fig.~\ref{f:work_overview}(b) and Fig.~\ref{f:work_overview}(c), respectively. For traditional memory cells, we first read weight $w_i$ from the corresponding memory cells. Afterward, the output $w_i$ is multiplied by the activation $x_i$ using a multiplier. Finally, we sum all products $x_i\cdot w_i$ from each of the multipliers together either by a single adder sequentially, or use a tree of adders to perform the sum in parallel. To achieve the same computation using EMT-based in-memory computing, we just need to connect the output of each memory cell to the same port. Physically, the sum of all the memory cell outputs $\sum x_i\cdot w_i$ can be obtained from that port directly. This is also referred to as a {\em current sum}.

\subsection{Challenges of Analog In-memory Deep Learning}

The energy consumption of a EMT cell is much less than that of traditional memory cells when they execute the MAC operation. There are important differences. In traditional memory cells, energy consumption is not related to the weight value that is stored. In analog EMT, it is proportional to the weight value~\cite{wang2020ncpower}, as shown in Fig.~\ref{f:basic}(a). We use a parameter $\rho$ to denote this energy coefficient. This parameter is tunable. We can use this parameter to optimize its energy consumption. Theoretically, a small coefficient $\rho$ can help us  improve the energy efficiency of models.

A big challenge in using the merging memory technology is that regardless of the actual technology, their memory cells do not output stable results~\cite{raghavan2013rtn}. Physically, each memory cell has multiple states, and it changes its state with time randomly. Whenever we read the memory cell, it can be in any of the states. At $l$-th state, the weight value read from the memory cell is $r_l(w,\rho)$, where $w$ is the pre-stored weight and $\rho$ is the energy coefficient. Given an input $x$, the output data becomes $x\cdot r_l(w,\rho)$ instead of $x\cdot w$. In Fig.~\ref{f:basic}(b), we show an example of a memory cell with two states. In this example, each state has a $50$\% probability to show. If the store weight is $w$, and the energy coefficient is $\rho$, the output result can be either $x\cdot r_0(w,\rho)$ or $x\cdot r_1(w,\rho)$ depending on its state, where $x$ is the input activation. The bottom three sub-graphs in Fig.~\ref{f:basic}(b) corresponds to memory cells with three different energy coefficients $\rho$.

The fluctuation shown in Fig.~\ref{f:basic}(b) is a simplified example. In practical EMT cells, the number of fluctuation states and the probability of each state are more complicated. There are many works in the literature studying these fluctuations~\cite{gong2018signal}.
For deep learning, such a phenomenon can cause a non-negligible accuracy drop. It is the biggest challenge that limits the application of in-memory computing. As we can see from Fig.~\ref{f:basic}(b), the fluctuation amplitude, defined as the average distance among $r_l(w,\rho)$, is related to the energy coefficient $\rho$. Theoretically, a higher $\rho$ will result in less fluctuation of the weight and thus higher model accuracy, but it also means higher energy consumption of memory cells. This trade off is the focus of this paper.

\subsection{Incompatibility of Traditional Training Method}
\label{s:risk}

\begin{figure}[!t]
  \centering
  \includegraphics[width=3.5in] {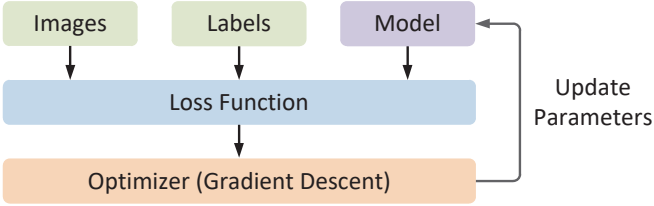}
  \caption{The training method for traditional deep learning. The dataset (green blocks) includes images and labels.}
  \label{f:overview_trad}
\end{figure}

Fig.~\ref{f:overview_trad} shows the standard training process in deep learning. The loss function takes the dataset and the model and generates a measure of the distance between the current parameter values and their optimal values. The optimizer uses gradient descent to reduce this distance by updating the parameters. This process would iterate for many epochs until the optimizer can find an optimal set of parameters. We define $\mathbf{X}\in \mathcal{X} $ as the image data and $\mathbf{Y}\in \mathcal{Y} $ as the label data. Here $\mathcal{X}$ and $\mathcal{Y}$ denote the spaces for images $\mathbf{X}$ and labels $\mathbf{Y}$. For simplicity, we express a one-layer neural network model in Equation~(\ref{e:whole}), where the weight $\mathbf{W}$ and the bias $\mathbf{B}$ are both trainable parameters.
\begin{equation}
\mathbf{Y}  = f(\mathbf{X}) = \mathbf{W}\mathbf{X}+\mathbf{B} 
\label{e:whole}
\end{equation}
Let's define the function class $\mathcal{F}\subset \mathcal{Y}^\mathcal{X}$ as the search space of function $f$. Let $ \mathcal{Z} = \mathcal{X} \times \mathcal{Y}$ be the combination of $\mathcal{X}$ and $\mathcal{Y}$, and $\textbf{Z}=(\textbf{X},\textbf{Y})$ be the combination of $\textbf{X}$ and $\textbf{Y}$. Define $\mathfrak{D}$ as the unknown distribution of data $\textbf{Z}$ on space $\mathcal{Z}$. Given the above definitions, the loss function can be expressed as $\mathcal{L}:\mathcal{F} \times \mathcal{Z} \rightarrow \mathbb{R}$. It is the mapping from the combination of $\mathcal{F}$ and $\mathcal{Z}$ to the real number space $\mathbb{R}$. Theoretically, the training process of the neural network is to find the optimal function $f$ from the function class $\mathcal{F}$ that can minimize the risk, i.e., the expectation of the loss function. We express it in Equation~(\ref{e:risk_old}). 
\begin{equation}
R[{f}]=\mathbb{E}_{{z}\sim {\mathfrak{D}}}\big[\mathcal{L}({f},{z})\big]
\label{e:risk_old}
\end{equation}

The difficulty in solving this problem is that the distribution $\mathfrak{D}$ is unknown. What we have is just the dataset $\{\textbf{Z}_1,\cdots,\textbf{Z}_N\}$, which are {\em independent and identically distributed} (i.i.d.) samples from the distribution $\mathfrak{D}$. Alternatively, the traditional training process of the neural network is to find the optimal model $f$ from the function class $\mathcal{F}$ that can minimize the empirical risk, i.e., the average of loss functions on the sampled dataset. We express it in Equation~(\ref{e:risk_oldd_sample}).
\begin{equation}
R_s[{f}]=\frac{1}{N}\sum_{i=1}^{N}{\mathcal{L}({f},{z_i})}
\label{e:risk_oldd_sample}
\end{equation}
The distance between the risk and the empirical risk is called the {\em generalization error} $\epsilon$, expressed in Equation~(\ref{e:risk_old_delta}). The crucial problem of the training process is to make sure that the generalization error $\epsilon$ can be bounded. 
\begin{equation}
\epsilon= R[{f}] - R_s[{f}]
\label{e:risk_old_delta}
\end{equation}

Researchers have since solved this math problem for traditional deep learning models. Many theoretical studies have shown that neural network optimizers, such as {\em stochastic gradient descent} (SGD)~\cite{bottou2012stochastic} or {\em adaptive moment estimation} (Adam)~\cite{zhang2018improved} can efficiently find the optimized function $f$. However, for in-memory deep learning models, because of the fluctuation of weight matrix $\mathbf{\widetilde{W}}$, the sampled function $f$ becomes $\widetilde{f}$ and the distance of risks becomes $\widetilde{\epsilon}$, as expressed in Equation~(\ref{e:risk_old_delta_2}). Therefore, the traditional training method no longer works for this problem. We need a new training method that is suitable for in-memory deep learning models.
\begin{equation}
\widetilde{\epsilon} = R[{f}] - R_s[\widetilde{f}]
\label{e:risk_old_delta_2}
\end{equation}

\begin{figure}[!t]
  \centering
  \includegraphics[width=3.5in] {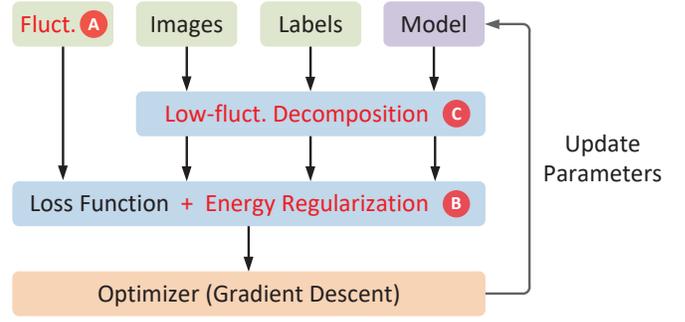}
  \caption{The proposed training method for in-memory deep learning. The dataset (green blocks) includes extra fluctuation data. We mark the proposed three optimization techniques in red color.}
  \label{f:overview}
\end{figure}

\begin{figure*}[!t]
  \centering
  \includegraphics[width=7in] {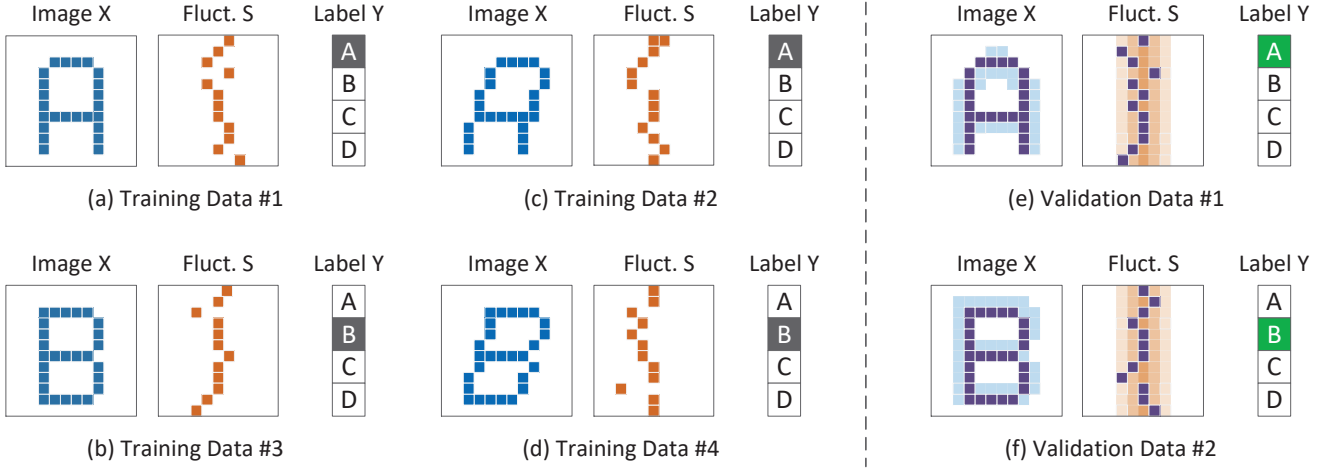}
  \caption{(a)-(f) Visualization of the device-enhanced datasets. Pixels forming letters \texttt{A} and \texttt{B} denote the image data labeled by class A and class B, respectively. The blurred pixels in (e) and (f) indicate the distribution of images $\mathbf{X}$ and fluctuation $\mathbf{S}$.}
  \label{f:process_path1}
\end{figure*}

\begin{figure*}[!t]
  \centering
  \includegraphics[width=7in] {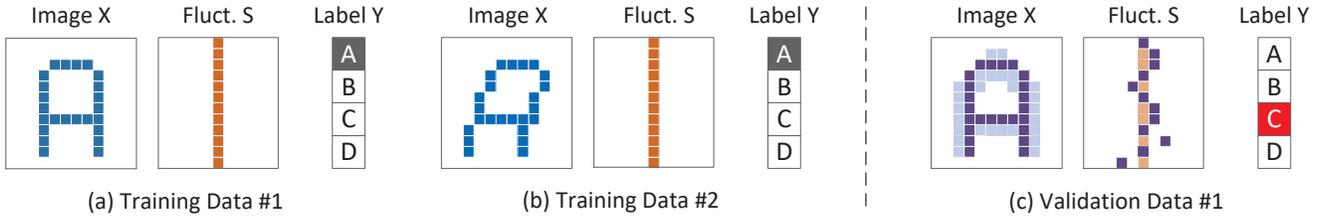}
  \caption{(a)-(c) Visualization of the traditional datasets. The orange straight lines in (a) and (b) indicate the absence of device information.}
  \label{f:process_path2}
\end{figure*}

\section{Optimizing for In-memory Deep Learning}
\label{s4}

Our training method for in-memory computing can effectively improve the model accuracy, with improved energy efficiency. Fig.~\ref{f:overview} shows the overview picture. Essentially, we propose three optimization techniques and integrate them into the training process. The first technique is called {\em device-enhanced datasets}. This technique integrates device information as additional data into the dataset. This will make the model more robust to the fluctuations of the device. The second technique is called {\em energy regularization} which we add a new regularization term into the loss function that makes the optimizer reduce the energy consumption automatically. The third technique is called {\em low-fluctuation decomposition} by which we decompose the computation involved into several time steps. This decomposition achieves high model accuracy and energy efficiency. We shall now give the mathematical basis of these three techniques.

\subsection{Device-Enhanced Dataset}

Our first optimization technique is to enhance the dataset with device information. In addition to the regular image data $\mathbf{X}$ and the label data $\mathbf{Y}$, our enhanced dataset has another source of data, the fluctuation $\mathbf{S}$, which reflects the random behavior of memory cells. Fig.~\ref{f:process_path1} shows an visualized example of $\mathbf{X}$, $\mathbf{Y}$ and $\mathbf{S}$, with four training data (Fig.~\ref{f:process_path1}(a))-(Fig.~\ref{f:process_path1}(d)). They can be classified as either letter \texttt{A} or letter \texttt{B}. Images in the same class can have different variants. Take the letter \texttt{A} for instance. It can be in any font, either normal (Fig.~\ref{f:process_path1}(a)) or italic (Fig.~\ref{f:process_path1}(c)). No matter what the variant is, its pixels must follow $\mathfrak{D}_\texttt{A}$, the distribution for class \texttt{A}. After training, we can accurately classify any image that belongs to class \texttt{A} as long as its pixels follow the distribution $\mathfrak{D}_A$ (Fig.~\ref{f:process_path1}(e)). This truth also holds for images of letter \texttt{B}, whose pixels follow distribution $\mathfrak{D}_\texttt{B}$. The fluctuation data $\mathbf{S}$ reflects the random states of memory cells. In our visualized example, the pixels indicate the states of the memory cells. The patterns for fluctuation $\mathbf{S}$ also follow a certain distribution $\mathfrak{R}$, which can be learned during training. Using this enhanced dataset, the model can make correct predictions for in-memory deep learning because now it becomes aware of the fluctuations (Fig.~\ref{f:process_path1}(e))-(Fig.~\ref{f:process_path1}(f)).

As we integrate the fluctuation device data into the dataset, the model will not overfit during training. In Fig.~\ref{f:process_path2}, we visualized an example of training using only the dataset $\mathbf{X}$ and do not consider the device information. All pixels in the data $\mathbf{S}$ are in the center of the matrix, indicating the absence of device information (Fig.~\ref{f:process_path2}(a))-(Fig.~\ref{f:process_path2}(b)). During training, the model will overfit this static data $\mathbf{S}$ because it does not have any variant. As we can see from Fig.~\ref{f:process_path2}(c), the distribution learned by the model is only a straight line (orange), which is different from the real fluctuation of memory cells (purple). Therefore, the model will mis-classify the images. On the other hand, if we include the fluctuation data $\mathbf{S}$, the overfit can be avoided. As we can see from Fig.~\ref{f:process_path2}(e) and Fig.~\ref{f:process_path2}(f), the fluctuation of memory cells (purple) will follow the learned distribution (orange) so that the model can accurately classify the images.

We developed a method to integrate the fluctuation data $\mathbf{S}$ into the training process. To simplified this problem, we first decompose the computation of neural network model (Equation~(\ref{e:whole})) into several sub-tasks. For example, each element $y_{ij}$ in the output matrix $\mathbf{Y}$ can be computed independently using Equation~(\ref{e:workload}). Here vector $\mathbf{w}_i$ is the $i$-th row in weight matrix $\mathbf{W}$, vector $\mathbf{x}_j$ is the $j$-th column in the input matrix $\mathbf{X}$, and $b_{ij}$ is the element in the bias matrix $\mathbf{B}$ at $i$-th row and $j$-th column.
\begin{equation}
y_{ij}  = \mathbf{w}_i\mathbf{x}_j+b_{ij}
\label{e:workload}
\end{equation}
As we showed in Fig.~\ref{f:basic}, the weight we read from the EMT cell is unpredictable. Physically, the memory cell changes its status randomly, and the exact output value depends on the state of the memory cell when we are reading it. We denote $w_{ik}$ be the $k$-th element in vector $\mathbf{w}_i$. $\widetilde{w}_{ik}(j)$ is the sampled data when we read $w_{ik}$ from the memory cell, and multiply it with the input vector $\mathbf{x}_j$. Mathematically, we can express $\widetilde{w}_{ik}(j)$ as a polynomial, as shown in Equation~(\ref{e:combination}).
\begin{equation}
\widetilde{w}_{ik}(j) = \sum_{l=1}^m{r_l(w_{ik}, \rho)\cdot s_{ijkl}} 
\label{e:combination}
\end{equation}
We use $r_l(w_{ik},\rho)$ to denote the weight retrieved when the memory cell is at $l$-th state. It can be considered as a function of the pre-stored weight $w_{ik}$ and energy coefficient $\rho$. At any moment, each memory cell can only be in one state. Hence, we use a one-hot encoded vector [$s_{ijkl}$]$_{1\leq l\leq m}$ to indicate the state of the memory cell when $w_{ik}$ is sampled. The value of $s_{ijkl}$ is shown in Equation~(\ref{e:coeff}). For given indexes $i$, $j$, and $k$, if the corresponding memory cell is at $l_{0}$-th state, only $s_{ijkl_0}$ equals $1$ and all the other coefficients equal $0$.
\begin{equation}
s_{ijkl}= 
\begin{cases}
    1  ~& \text{if } l = l_{0}\\
    0  ~& \text{if } l \neq l_{0}
\end{cases}
\label{e:coeff}
\end{equation}

As we can see from Equation~(\ref{e:combination}), the sampled weight $\widetilde{w}_{ij}$ from the memory cell consists of two parts:

\begin{itemize}
\item The {{\textit{deterministic}}} parameter $r_l(w_{ik},\rho)$ is a function indicating the returned value from the memory cell storing weight $w_{ik}$ for a memory cell that is in the $l$-th state. We denote the matrix $[r_l(w_{ik}),\rho]$, as $\mathbf{r}(\mathbf{w}_i,\rho)$, shown in Equation~(\ref{e:matrix1}). $\mathbf{r}(\mathbf{w_i},\rho)$ can be considered a function of the weight vector $\mathbf{{w}}_i$ and the energy coefficient $\rho$.

\item The {{\textit{stochastic}}} parameter $s_{ijkl}$ is a random coefficient indicating whether the memory cell storing weight $w_{ik}$ is at $l$-th state when it is sampled and multiplied with the input vector $\mathbf{x}_j$. We denote matrix $[s_{ijkl}]$ as $\mathbf{S}_{ij}$, shown in Equation~(\ref{e:matrix2}). $\mathbf{S}_{ij}$ can be considered as a part of fluctuation data $\mathbf{S}$. 
\end{itemize}
\begin{align}
\mathbf{r}(\mathbf{w}_i,\rho) =& 
\begin{bmatrix}
r_1(w_{i1},\rho) &r_1(w_{i2},\rho) &\cdots &r_{1}(w_{in},\rho) \\
r_2(w_{i1},\rho) &r_2(w_{i2},\rho) &\cdots &r_{2}(w_{1n},\rho) \\
\vdots  &\vdots  &\ddots &\vdots  \\
r_m(w_{i1},\rho) &r_m(w_{i2},\rho) &\cdots &r_{m}(w_{in},\rho)
\end{bmatrix}\label{e:matrix1}\\
\mathbf{S}_{ij} = &
\begin{bmatrix}
s_{ij11} &s_{ij21} &\cdots &s_{ijn1} \\
s_{ij12} &s_{ij22} &\cdots &s_{ijn2} \\
\vdots  &\vdots  &\ddots &\vdots  \\
s_{ij1m} &s_{ij2m} &\cdots &s_{ijnm}
\end{bmatrix}
\label{e:matrix2}
\end{align}
We can now integrate the fluctuation data $\mathbf{S}$ into the training process for in-memory deep learning. Each element $y_{ij}$ in the output matrix $\mathbf{Y}$ can be calculated using Equation~(\ref{e:sample}). $y_{ij}$ is a function of both the deterministic and stochastic parameters. Here $\mathbf{\widetilde{w}}_i$ refers to the sampled weight vector read from the memory cells. For simplicity, we assume the bias $b_{ij}$ is a deterministic parameter. In some cases, the bias is also fluctuating. We can use the same method to separate deterministic and stochastic parameters for $b_{ij}$.
\begin{equation}
y_{ij}  = \mathbf{\widetilde{w}}_i\mathbf{x}_j + b_{ij} =\mathbf{1}\big(\mathbf{r}(\mathbf{w}_i,\rho)\circ \mathbf{S}_{ij}\big)\mathbf{x}_j + b_{ij}
\label{e:sample}
\end{equation}
The $\circ$ operator between $r(\mathbf{w}_i, \rho)$ and $\mathbf{S_{ij}}$ is the {\em Hadamard product}, i.e., element-wise product. The unit vector $\mathbf{1}$ is expressed in Equation~(\ref{e:unit}). We use it to sum up the entire column of the target matrix. 
\begin{equation}
\mathbf{1} = 
{\small \underbrace{\begin{bmatrix}
1 &1  &\cdots &1 \\
\end{bmatrix}}_{m}}
\label{e:unit}
\end{equation}

\subsection{Energy Regularization}

\begin{figure}[!t]
  \centering
  \includegraphics[width=3.5in] {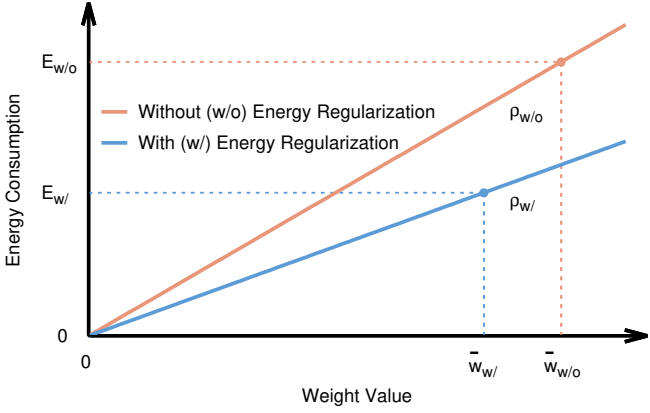}
  \caption{Energy-saving from the energy regularization term. It can decrease both the energy coefficient $\rho$ and the mean of weight values $\overline{w}$.}
  \label{f:energy_term}
\end{figure}

Our second optimization technique adds an energy regularization term into the loss functions during training. From Equation~(\ref{e:sample}) we can infer that the loss function of the model $\mathcal{L}_0$ is a function of weights $\mathbf{w}$ and energy coefficient $\rho$. The target of our optimization technique is to find the optimal energy coefficient $\rho$ that can improve both the model accuracy and energy efficiency. However, it is not an easy task. We prefer a smaller $\rho$ for higher energy efficiency. However, as we see in Fig.~\ref{f:basic}, the higher $\rho$ causes a larger fluctuation amplitude of the weights, which results in accuracy loss. On the other hand, if we choose a larger coefficient $\rho$, the model accuracy would be less affected by the weight fluctuation, but the energy consumption becomes larger.

Our new loss function $\mathcal{L}$ is expressed in Equation~(\ref{e:energy_reg}). The first term $\mathcal{L}_0$ is the original loss function of the model, and the second term represents the energy consumption of the model. $\lambda$ is a hyper-parameter indicating the significance of the energy regularization term. $\alpha_t$ is a constant indicating the number of reading operations from the memory cell 
storing weight $w_t$. The overall loss function $\mathcal{L}$ can be considered as a function of $\textbf{w}$ ($w_t$ is the $t$-th element of $\textbf{w}$) and $\rho$, which are both trainable parameters. We can use any popular optimizer (such as SGD optimizer~\cite{bottou2012stochastic} or Adam optimizer~\cite{zhang2018improved}) to search for the optimal weight $\mathbf{w}$ and energy coefficient $\rho$.
\begin{equation}
\mathcal{L}(\textbf{w},\rho)= \mathcal{L}_0(\textbf{w}, \rho) + \lambda \sum_{t}{\alpha_t \rho |w_t|}
\label{e:energy_reg}
\end{equation}

During training, gradient descent will minimize the loss function $\mathcal{L}$. After optimization, both $\rho$ and $w_t$ will become smaller. We show this process in Fig.~\ref{f:energy_term}. With the help of the energy regularization term, we can improve both the model accuracy and energy efficiency simultaneously.

\subsection{Low-fluctuation Decomposition}

\begin{figure}[!t]
  \centering
  \includegraphics[width=3.5in] {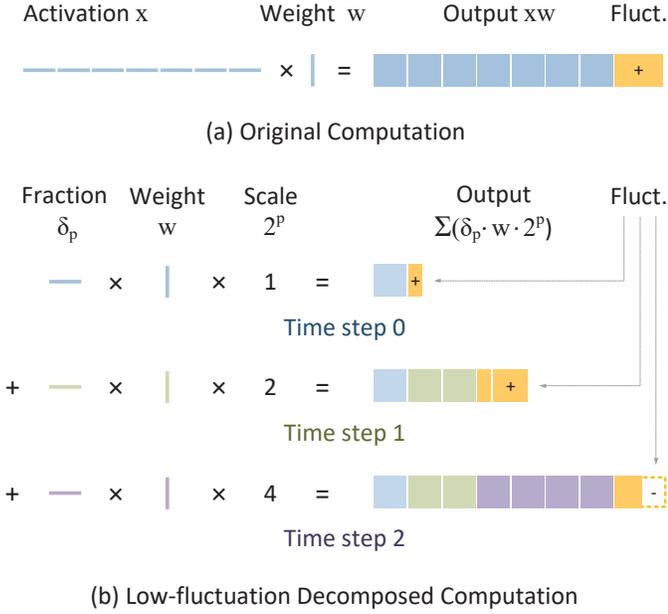}
  \caption{(a) Original computation mechanism; (b) low-fluctuation decomposed computation mechanism. The length of the bar denotes activation/weight value. The area of the square denotes their product. The solid/hollow yellow block represents positive/negative fluctuation.}
  \label{f:coding}
\end{figure}

The third optimization technique is to decompose the computation process into multiple time steps. We can visualize the computation involved in Fig.~\ref{f:coding}. The input activation $x$ and the weight $w$ equals the length of the horizontal bar and the vertical bar, respectively. The computation result got from the memory cell equals the area of the square, whose two edges have the same length as the horizontal bar and the vertical bar. In the example of original computing (Fig.~\ref{f:coding}(a)), the lengths of the horizontal bar and vertical bar are 
seven ($x=7$) and one ($w=1$), respectively. The area of the output square is thus seven ($x\cdot w=7$).

Theoretically, we can express any input $x$ as a polynomial, as shown in Equation~(\ref{e:decom}). Here the fraction bit $\delta_p$ is binary data, which equals either $0$ or $1$, and $2^p$ is the scaling factor for that term. For example, if the input equals $7$, we can decompose it into three parts: $1\cdot 2^0$, $1\cdot 2^1$, and $1\cdot 2^2$. 
\begin{equation}
x= \sum{(\delta_p\cdot 2^p)}
\label{e:decom}
\end{equation}
In our low-fluctuation decomposed computation mechanism (Fig.~\ref{f:coding}(b)), we read each memory cell in multiple time steps instead of once. As Equation~(\ref{e:accum}) shows, at each time step, we only input the fraction bit $\delta_p$ to the memory cell, obtain the value of $\delta_p\cdot w$, and then scale the output from the memory cell by the factor $2^p$. Finally, we sum up all the results from each time step. In this example,  we use three steps to process input seven ($x=7$). The scaling factor of each time step is $2^0$, $2^1$, and $2^2$, respectively. As the weight is one ($w=1$), the final accumulated result is seven ($x\cdot w=7$), the same result as the original computing mechanism.
\begin{equation}
x\cdot w= \sum{(\delta_p\cdot w \cdot 2^p)}
\label{e:accum}
\end{equation}

As the name indicates, our low-fluctuation decomposition can alleviate the fluctuations of the memory cell effectively. We can explain this using Fig.~\ref{f:coding}, where we show the fluctuation amplitude in the yellow blocks. The block and hollow block denote positive and negative fluctuation amplitudes, respectively. As we can see from Fig.~\ref{f:coding}(b), using the decomposed computation mechanism,  the negative fluctuation amplitude (hollow block) at the third time step can partially average out the positive fluctuation amplitude at the second time step (solid block). Statistically, the accumulated fluctuations from the decomposed computation mechanism have a lower standard deviation than that of the original computation mechanism.

We can mathematically compare their standard deviations. Equation~(\ref{e:o_trad}) shows the standard deviation of the original computation mechanism, where $O_{\text{ori}}$ is the original output, and $\sigma(w)$ is the standard deviation of $w$ when we read it from the memory cell. Equation~(\ref{e:o_new}) shows the standard deviation of our low-fluctuation decomposed computation mechanism, where $O_{\text{new}}$ is the new output, and $w(p)$ is the weight sampled from the memory cell at $p$-th time step. Since reading memory cells can be considered as independent events, we have $\sigma(w(p)) = \sigma(w)$.
\begin{align}
\ \sigma(O_{\text{ori}}) =\sigma(x\cdot w) = \sum{2^{p}\delta_p}\cdot \sigma(w)
\label{e:o_trad}
\end{align}
\begin{equation}
\begin{aligned}
\sigma(O_{\text{new}})= \sigma\big(\sum{2^{p}\delta_pw(p)}\big) = &\sqrt{\sum{2^{2p}\delta^2_p\sigma^2\big (w(p)\big )}} \\
=& \sqrt{\sum{2^{2p}\delta^2_p}}\cdot \sigma(w)\\
<& \sqrt{(\sum{2^{p}\delta_p})^2}\cdot \sigma(w)
\label{e:o_new}
\end{aligned}
\end{equation}
We can infer from Equation~(\ref{e:o_trad}) and~(\ref{e:o_new}) that our decomposed computation result has a lower standard deviation than the original computation result, leading to high model accuracy (Equation~(\ref{e:o_compare})).
\begin{align}
\sigma(O_{\text{new}}) < \sigma(O_{ori})  
\label{e:o_compare}
\end{align}

Our low-fluctuation decomposition can also improve energy efficiency. To prove this, we express the energy consumption of the two computation mechanisms in Equation~(\ref{e:e_list}). From the equation, we can infer that our decomposed computation mechanism consumes less energy than the original one (Equation~(\ref{e:e_compare})).
\begin{align}
&E(O_{\text{ori}}) = \rho\cdot x \ \ \ \ \ \  E(O_{\text{new}}) = \rho \sum \delta_p \label{e:e_list} \\
&E(O_{\text{new}}) < E(O_{\text{ori}})  
\label{e:e_compare}
\end{align}

\begin{figure*}[!t]
  \centering
  \includegraphics[width=7in] {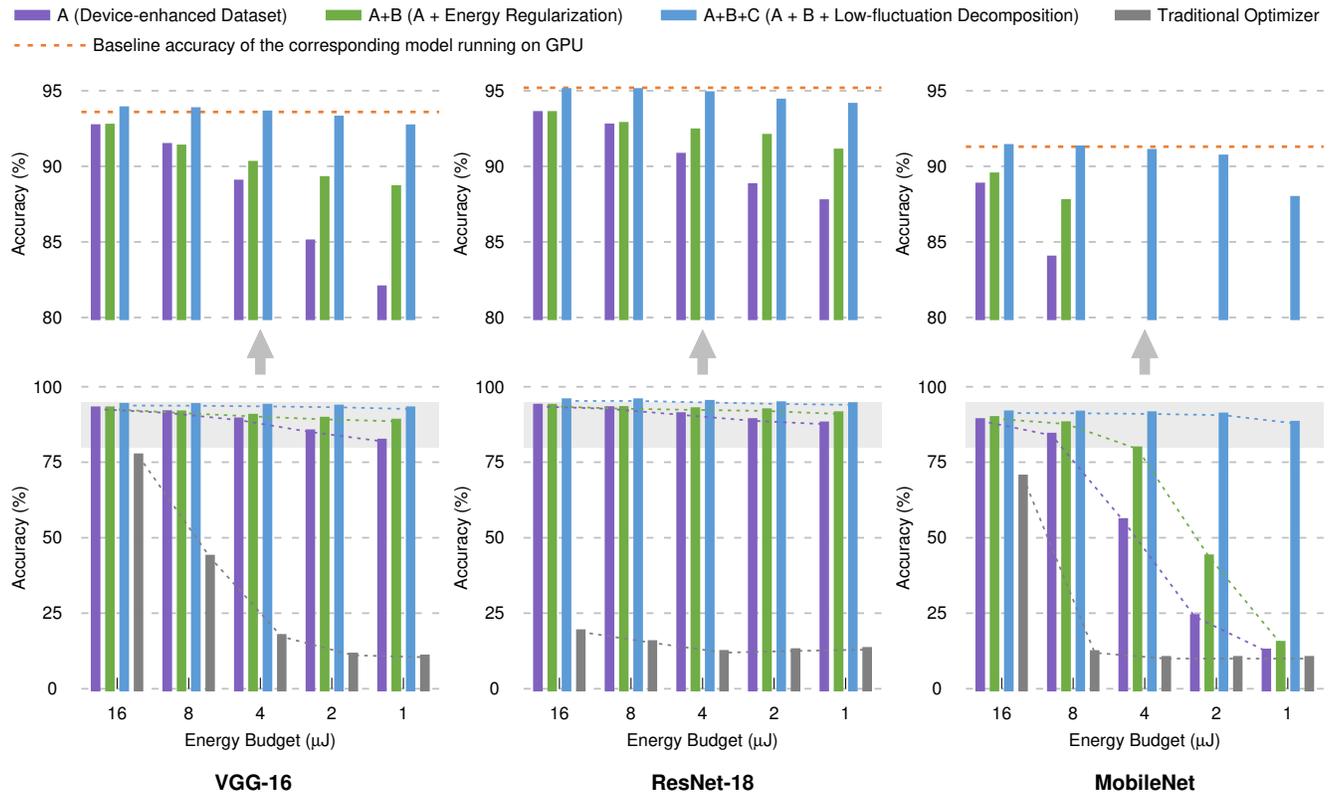}
  \caption{The comparison between our proposed optimization solutions and the traditional optimizer. We test models on the CIFAR-10 dataset. The first row of sub-figures shows the accuracies in a zoomed range, and the second row of sub-figures shows accuracies in the full range.}
  \label{f:performance}
\end{figure*}

\subsection{Convergence of the Training Method}

We shall now mathematically prove the convergence of our training method. 
Equation~(\ref{e:whole2}) shows the basic relationship between image data $\mathbf{X}$ and label data $\mathbf{Y}$ in the traditional deep learning network. The output matrix $\mathbf{Y}\in \mathcal{Y}$ is a function $f$ of the input matrix $\mathbf{X}\in \mathcal{X}$. Here $\mathcal{X}$ and $\mathcal{Y}$ denote the space of $\mathbf{X}$ and $\mathbf{Y}$, respectively. 
\begin{equation}
\mathbf{Y}  = f(\mathbf{X})
\label{e:whole2}
\end{equation}
For in-memory deep learning applications, the computation becomes unpredictable because of weight fluctuation. As can be seen from Equation~(\ref{e:sample}), output $y_{ij}$ is a function of both the input $\mathbf{x}_j$ and the fluctuation data $\mathbf{S}_{ij}$. To generalize this, the output matrix $\mathbf{Y}\in \mathcal{Y}$ can be defined a function $\widetilde{f}$ of the input matrix $\mathbf{X}\in \mathcal{X}$ and the fluctuation data $\mathbf{S}\in \mathcal{S}$, shown in Equation~(\ref{e:PIM}). Here $\mathcal{S}$ denotes the space of $\mathbf{S}$.
\begin{equation}
\mathbf{Y} = \widetilde{f}(\mathbf{X, S})
\label{e:PIM}
\end{equation}
We define a new data $\widetilde{\mathbf{Z}}\in \widetilde{\mathcal{Z}}$ as the combination of images $\mathbf{X}$, labels $\mathbf{Y}$ and fluctuations $\mathbf{S}$.
Here the space $\widetilde{\mathcal{Z}}$ is the combination of $\mathcal{X}$, $\mathcal{Y}$, and  $\mathcal{S}$. Since  $\mathcal{Z}$ is the combination of $\mathcal{X}$ and $\mathcal{Y}$, we can express the space $\widetilde{\mathcal{Z}}$ as the combination of the spaces $\mathcal{Z}$ and $\mathcal{S}$. We show their relationships in Equation~(\ref{e:parameter1})-(\ref{e:parameter2}).
\begin{align}
\widetilde{\mathbf{Z}}=&(\mathbf{X}, \mathbf{Y}, \mathbf{S})\label{e:parameter1}\\
\widetilde{\mathcal{Z}}=&\mathcal{X}\times \mathcal{Y}\times\mathcal{S}= \mathcal{Z}\times \mathcal{S}
\label{e:parameter2}
\end{align}
Given the fact that $\mathcal{Z}$ follows distribution $\mathfrak{D}$, while $\mathcal{S}$ follows distribution $\mathfrak{R}$, we can infer that the space $\widetilde{\mathcal{Z}}$ (the combination of $\mathcal{Z}$ and $\mathcal{S}$) must follow a distribution $\widetilde{\mathfrak{D}}$ (the combination of $\mathfrak{D}$ and $\mathfrak{R}$). We show this in Equation~(\ref{e:new_dist}).
\begin{equation}
       \mathcal{Z}\sim\mathfrak{D}, \mathcal{S}\sim\mathfrak{R} 
\ \ \ \Rightarrow
\ \ \ \widetilde{\mathcal{Z}}\sim \widetilde{\mathfrak{D}}
\label{e:new_dist}
\end{equation}

Finally, our proposed training process for in-memory deep learning models can be concluded as follows: given a function $\widetilde{f}$ in the space $\widetilde{\mathcal{F}}$, and a loss function $\mathcal{L}:\widetilde{\mathcal{F}} \times \widetilde{\mathcal{Z}}\rightarrow \mathbb{R}$, we would like to find $\widetilde{f}\in \widetilde{\mathcal{F}}$ that can minimize the risk, \textit{i.e.}, the expectation of the loss function (Equation~(\ref{e:risk})).
\begin{equation}
R[\widetilde{f}]=\mathbb{E}_{\widetilde{z}\sim \widetilde{\mathfrak{D}}}\big[\mathcal{L}(\widetilde{f},\widetilde{z})\big]
\label{e:risk}
\end{equation}

Thus, we convert the convergence problem of the new training process for in-memory computing into the convergence problem for regular neural networks (see Section~\ref{s:risk}). Finding the optimal function is well studied, and we can use various existing optimizers, such as SGD or Adam, to find the optimal function $\widetilde{f}$.

\section{Experiments}
\label{s5}

We trained the models on the Pytorch platform and evaluate the energy consumption in an in-memory deep learning simulation platform~\cite{lee2019system}~\cite{wang2020ncpower}. To accelerate the training process, we start each experiment from a well-trained model with full-precision weights~\cite{pytorchzoo} and then fine-tune the model by applying our proposed optimizations. During fine-tuning, we quantize both the activations and weights. To form the device-enhance dataset, we fetch the images and labels from the regular datasets (such as CIFAR-10 and ImageNet) as data $\textbf{X}$ and data $\textbf{Y}$, respectively. The fluctuation data $\textbf{S}$ are obtained from state-of-the-art device models~\cite{ielmini2010resistance}. We use a workstation with an Nvidia 2080TI graphic card to train the model. For the CIFAR-10/ImageNet dataset, each experiment can finish in about an hour/an entire day.

We proposed three solutions, denoted as `A',  `A+B', and `A+B+C'. As shown in Fig.~\ref{f:overview}, the notations A, B, C stand for the device-enhanced dataset, energy regularization, and low-fluctuation decomposition, respectively. Solution A uses only the first technique; solution A+B applies the first two; solution A+B+C combines all three. We evaluate popular models including VGG-16, ResNet-18/34 and MobileNet. VGG-16 is a regular deep neural network with only $3\times 3$ kernels. ResNet-18/34 are popular models that achieve competitive accuracy by adding residual links between layers. MobileNet is a small-size model that achieves high efficiency due to its special depthwise layer. We compare our work with three state-of-the-art solutions: binarized encoding~\cite{zhu2019configurable}, weight scaling~\cite{ielmini2010resistance}, and fluctuation compensation~\cite{wan2020voltage} as described in detail in Section~\ref{s2}.

\begin{figure*}[!t]
  \centering
  \includegraphics[width=7in] {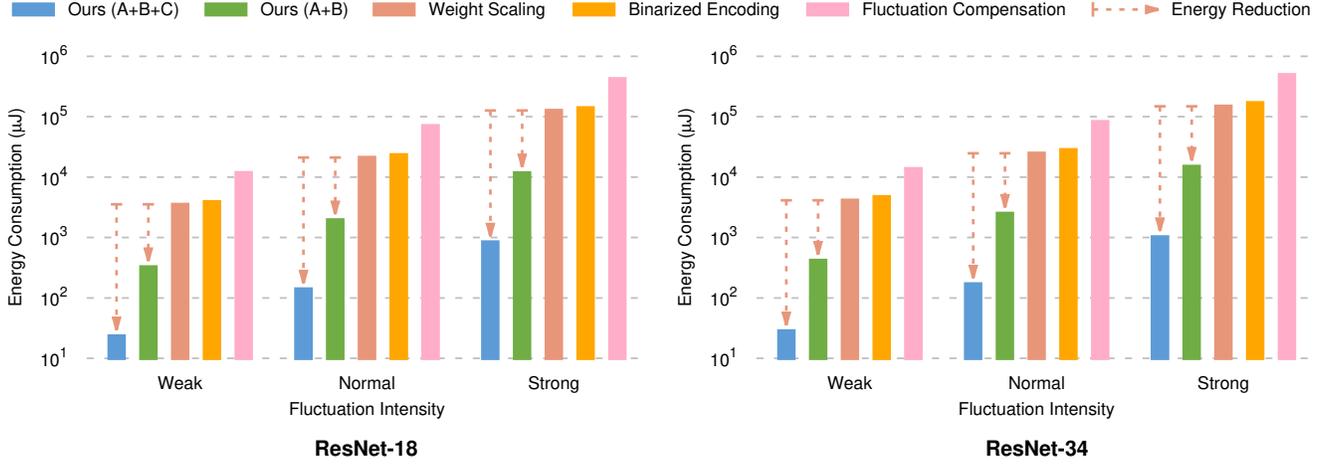}
  \caption{The energy comparison between our proposed solutions and the state-of-the-art, under three levels of fluctuation intensity. We test models on the ImageNet dataset. Both our solutions and the state-of-the-art are free to tune the energy coefficient $\rho$.}
  \label{f:rob}
\end{figure*}

\begin{figure*}[!t]
  \centering
  \includegraphics[width=7in] {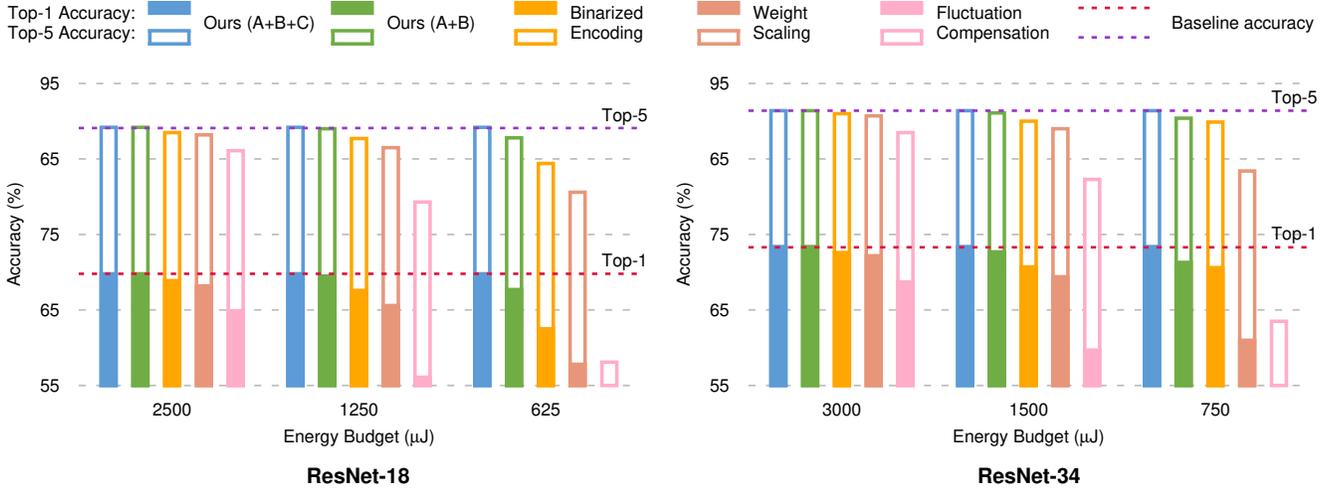}
  \caption{The accuracy comparison between our proposed solutions and the state-of-the-art. We test models on the ImageNet dataset. The solid bar and hollow bar denote the top-1 accuracy and top-5 accuracy, respectively. The dash line denotes baseline model accuracy on GPU.}
  \label{f:sota}
\end{figure*}

\subsection{Ablation Study of Proposed Techniques}

Models optimized by our proposed methods have much higher accuracy than the model trained by the traditional optimizer. In Fig.~\ref{f:performance}, we show the accuracy achieved by solution A, solution A+B, and solution A+B+C under different energy budgets. As a reference, we also give the model accuracy trained by the traditional optimizer. As we can see from the figure, at 16 $\mu$J energy budget, the accuracy of solution A+B+C is very close to baseline accuracy (shown as dashed lines in the top sub-figures). On the other hand, the traditional optimizer exhibits relatively low accuracy due to its unawareness of memory fluctuation.

We can see that when the energy budget is decreased, models trained using the traditional optimizer show a dramatic decrease in accuracy. On the contrary, our solution A+B+C can achieve high model accuracy even if we reduce the energy budget. Even just using Solutions A and A+B is enough to maintain a relatively high accuracy, only to be outperformed by Solution A+B+C. This observation further proves the effectiveness of our proposed three techniques on in-memory computing.

We can also see that under 16 $\mu$J energy consumption, the ResNet-18 trained by the traditional optimizer shows much lower accuracy than the VGG-16. By using our solution A+B+C, ResNet-18 can fully recover the accuracy and thus outperforms VGG-16. This experiment shows that MobileNet is not suitable for in-memory deep learning. Under the same energy budget, MobileNet shows lower accuracy than VGG-16 and ResNet-18. We attribute this to its depthwise layer. When we compute a regular convolution layer, the system reads hundreds of memory cells at once. However, to process the depthwise layer, it only read nine memory cells at once. Therefore, a large portion of the energy is consumed in the peripheral circuits, causing a significant amount of energy overhead.

\setlength{\tabcolsep}{4pt}

\renewcommand{\arraystretch}{1.26}

\begin{table*}[!t]
\caption{Comparison between our solutions and the state-of-the-art on the CIFAR-10 dataset}
\resizebox{1.00\linewidth}{!}{
\begin{tabular}{l  c | c    c  c | c c c   | c c  c}
\hline	
\hline	
& &\multicolumn{3}{c|}{0\% accuracy drop}&\multicolumn{3}{c|}{1\% accuracy drop} &\multicolumn{3}{c}{2\% accuracy drop}\\
\hline											
VGG-16 (93.6\% Acc.)	&Ref.	&Energy ($\mu$J)	&\#Cells	&Delay ($\mu$S)	&Energy ($\mu$J)	&\#Cells	&Delay ($\mu$S)	&Energy ($\mu$J)	&\#Cells	&Delay ($\mu$S)	\\
\hline											
Binarized Encoding	&\cite{zhu2019configurable}	&378	&74M	&2.8	&135	&74M	&2.8	&94	&74M	&2.8	\\
Weight Scaling	&\cite{ielmini2010resistance}	&444	&15M	&2.8	&78	&15M	&2.8	&49	&15M	&2.8	\\
Fluctuation Compensation	&\cite{wan2020voltage}	&1091	&15M	&14	&157	&15M	&14	&82	&15M	&14	\\
\textbf{Ours (A+B)}	&	&\textbf{36}	&\textbf{15M}	&\textbf{2.8}	&\textbf{16}	&\textbf{15M}	&\textbf{2.8}	&\textbf{11}	&\textbf{15M}	&\textbf{2.8}	\\
\textbf{Ours (A+B+C)}	&	&\textbf{4.1}	&\textbf{15M}	&\textbf{14}	&\textbf{1.0}	&\textbf{15M}	&\textbf{14}	&\textbf{0.5}	&\textbf{15M}	&\textbf{14}	\\
\hline											
ResNet-18 (95.2\% Acc.)	&Ref.	&Energy ($\mu$J)	&\#Cells	&Delay ($\mu$S)	&Energy ($\mu$J)	&\#Cells	&Delay ($\mu$S)	&Energy ($\mu$J)	&\#Cells	&Delay ($\mu$S)	\\
\hline											
Binarized Encoding	&\cite{zhu2019configurable}	&876	&56M	&6.8	&389	&56M	&6.8	&286	&56M	&6.8	\\
Weight Scaling	&\cite{ielmini2010resistance}	&1127	&11M	&6.8	&209	&11M	&6.8	&158	&11M	&6.8	\\
Fluctuation Compensation	&\cite{wan2020voltage}	&2217	&11M	&34	&474	&11M	&34	&347	&11M	&34	\\
\textbf{Ours (A+B)}	&	&\textbf{83}	&\textbf{11M}	&\textbf{6.8}	&\textbf{22}	&\textbf{11M}	&\textbf{6.8}	&\textbf{10}	&\textbf{11M}	&\textbf{6.8}	\\
\textbf{Ours (A+B+C)}	&	&\textbf{6.9}	&\textbf{11M}	&\textbf{34}	&\textbf{1.1}	&\textbf{11M}	&\textbf{34}	&\textbf{0.7}	&\textbf{11M}	&\textbf{34}	\\
\hline											
MobileNet (91.3\% Acc.)	&Ref.	&Energy ($\mu$J)	&\#Cells	&Delay ($\mu$S)	&Energy ($\mu$J)	&\#Cells	&Delay ($\mu$S)	&Energy ($\mu$J)	&\#Cells	&Delay ($\mu$S)	\\
\hline											
Binarized Encoding	&\cite{zhu2019configurable}	&392	&16M	&4.6	&81	&16M	&4.6	&62	&16M	&4.6	\\
Weight Scaling	&\cite{ielmini2010resistance}	&232	&3.2M	&4.6	&57	&3.2M	&4.6	&42	&3.2M	&4.6	\\
Fluctuation Compensation	&\cite{wan2020voltage}	&659	&3.2M	&23	&126	&3.2M	&23	&91	&3.2M	&23	\\
\textbf{Ours (A+B)}	&	&\textbf{75}	&\textbf{3.2M}	&\textbf{4.6}	&\textbf{23}	&\textbf{3.2M}	&\textbf{4.6}	&\textbf{13}	&\textbf{3.2M}	&\textbf{4.6}	\\
\textbf{Ours (A+B+C)}	&	&\textbf{12.2}	&\textbf{3.2M}	&\textbf{23}	&\textbf{1.8}	&\textbf{3.2M}	&\textbf{23}	&\textbf{1.3}	&\textbf{3.2M}	&\textbf{23}	\\
\hline											
\hline																																																
\label{t:sota1}
\end{tabular}
}
\end{table*}

\begin{table*}[!t]
\caption{Comparison between our solutions and the state-of-the-art on the ImageNet dataset}
\resizebox{1.00\linewidth}{!}{
\begin{tabular}{l  c | c    c  c | c c c   | c c  c}
\hline	
\hline	
& &\multicolumn{3}{c|}{0\% accuracy drop}&\multicolumn{3}{c|}{1\% accuracy drop} &\multicolumn{3}{c}{2\% accuracy drop}\\
\hline											
ResNet-18(69.8\% Acc.)	&Ref.	&Energy ($\mu$J)	&\#Cells	&Delay ($\mu$S)	&Energy ($\mu$J)	&\#Cells	&Delay ($\mu$S)	&Energy ($\mu$J)	&\#Cells	&Delay ($\mu$S)	\\
\hline											
Binarized Encoding	&\cite{zhu2019configurable}	&23k (\textcolor{red}{-0.4\%})	&58M	&151	&2338	&58M	&151	&1336	&58M	&151	\\
Weight Scaling	&\cite{ielmini2010resistance}	&21k (\textcolor{red}{-0.3\%})	&12M	&151	&3544	&12M	&151	&1933	&12M	&151	\\
Fluctuation Compensation	&\cite{wan2020voltage}	&71k (\textcolor{red}{-0.3\%})	&12M	&756	&8505	&12M	&756	&4725	&12M	&756	\\
\textbf{Ours (A+B)}	&	&\textbf{1951}	&\textbf{12M}	&\textbf{151}	&\textbf{897}	&\textbf{12M}	&\textbf{151}	&\textbf{659}	&\textbf{12M}	&\textbf{151}	\\
\textbf{Ours (A+B+C)}	&	&\textbf{142}	&\textbf{12M}	&\textbf{756}	&\textbf{71}	&\textbf{12M}	&\textbf{756}	&\textbf{54}	&\textbf{12M}	&\textbf{756}	\\
\hline											
ResNet-34 (73.3\% Acc.)	&Ref.	&Energy ($\mu$J)	&\#Cells	&Delay ($\mu$S)	&Energy ($\mu$J)	&\#Cells	&Delay ($\mu$S)	&Energy ($\mu$J)	&\#Cells	&Delay ($\mu$S)	\\
\hline											
Binarized Encoding	&\cite{zhu2019configurable}	&28k (\textcolor{red}{-0.2\%})	&109M	&207	&2844	&109M	&207	&1778	&109M	&207	\\
Weight Scaling	&\cite{ielmini2010resistance}	&25k (\textcolor{red}{-0.1\%})	&22M	&207	&3302	&22M	&207	&2154	&22M	&207	\\
Fluctuation Compensation	&\cite{wan2020voltage}	&83k (\textcolor{red}{-0.1\%})	&22M	&1033	&7990	&22M	&1033	&4857	&22M	&1033	\\
\textbf{Ours (A+B)}	&	&\textbf{2496}	&\textbf{22M}	&\textbf{207}	&\textbf{1044}	&\textbf{22M}	&\textbf{207}	&\textbf{729}	&\textbf{22M}	&\textbf{207}	\\
\textbf{Ours (A+B+C)}	&	&\textbf{168}	&\textbf{22M}	&\textbf{1033}	&\textbf{90}	&\textbf{22M}	&\textbf{1033}	&\textbf{62}	&\textbf{22M}	&\textbf{1033}	\\
\hline											
\hline											
\label{t:sota2}
\end{tabular}
}
\end{table*}

\subsection{Robustness to Different Devices}

Today, academia and industry have developed various types of  EMT cells, which have different levels of fluctuation intensity. Hence, it is necessary to prove the robustness of our solutions under any level of fluctuation intensity. In Fig.~\ref{f:rob}, we test our solutions and the state-of-the-art under three intensity levels~\cite{raghavan2013microscopic}:  weak,  normal, and strong. The experiment is conducted on the ImageNet dataset using two ResNet models. All solutions, including the state-of-the-art, are free to tune the energy coefficient $\rho$. We compare the energy consumption when the model achieves its maximum accuracy. Noted that on the ImageNet dataset, our solutions can achieve the same accuracy as the baseline model running on GPU, where the state-of-the-art cannot.

The results show the robustness of our solutions. At any level of fluctuation intensity, our solution has almost the same performance on energy reduction. When fluctuation intensity is increased, both ours and the state-of-the-art solutions prefer a higher energy coefficient $\rho$ to maximize the model accuracy, resulting in a higher energy consumption. However, our solutions still outperform the state-of-the-art. Our solutions A+B and A+B+C shows one and two orders of magnitude energy reduction, respectively. From this point onwards, we shall assume the fluctuation intensity to be normal. Results under any other intensity level follow a similar trend.

\subsection{Verification of the Optimization Solutions}

In Fig.~\ref{f:sota}, we verified our optimization solution by testing two ResNet models on the ImageNet dataset. The dashed line in the figure shows the baseline accuracy. We defined it as the highest accuracy we can achieve on GPU. We also list the accuracy of the state-of-the-art for comparison. Among all, our solution A+B+C has the highest top-1 and top-5 accuracies, which are the same as the baseline accuracies. The accuracy of solution A+B also shows higher accuracy than the state-of-the-art. We can observe a small accuracy loss under a smaller energy budget. By contrast, models optimized by the state-of-the-art have significant accuracy losses. For example, we can see at least 0.9\% and 0.8\% top-1 accuracy losses of the ResNet-18 and ResNet-34, respectively. We can also see that the ResNet-18 on the ImageNet dataset consumes more energy than the same model on the CIFAR-10 dataset. It is because ImageNet has a larger image size than CIFAR-10.

\subsection{Holistic Comparison with the State-of-the-Art}

Our proposed solutions have better performance not only on energy reduction but also reduce cost and latencies. In Table~\ref{t:sota1} and Table~\ref{t:sota2}, we give a holistic comparison of our solutions with the state-of-the-art on energy consumption, the number of cells, and latency, under the same accuracy loss. We test various models on the CIFAR-10 and the ImageNet datasets. From the tables, our method has the lowest energy consumption, the least number of cells, and the shorted latency. Specifically, solution A+B shows one order of magnitude improvement in energy consumption, and solution A+B+C can achieve two orders of magnitude improvement, compared with the state-of-the-art. However, one limitation of Solution A+B+C is that its latency is longer because the low-fluctuation decomposed computation takes time to accumulate the results. The trade-off between energy consumption and latency depends on the specific application. Therefore, we list the results of solution A+B and solution A+B+C. Developers can choose one of them based on their demands.

Another observation is that our solutions are the only ones that can fully recover the accuracy loss on the ImageNet datasets. As can be seen from the third column of Table~\ref{t:sota2}, all the state-of-the-art solutions are unable attain no accuracy loss (we mark their actual accuracy drops in red text). In a neural network, some parameters are vital to achieving high model accuracy. However, the state-of-the-art usually applies a general optimization rule to all the model parameters. After the optimization, those important parameters still have relatively large fluctuations, which constraints the recovery of model accuracy. Unlike the state-of-the-art, our solutions can automatically identify those significant parameters and minimize their fluctuations. 

\section{Conclusion}
\label{s6}

In-memory deep learning has a promising future in the AI industry because of its high energy efficiency over traditional deep learning. This is even more so if the potential of emerging memory technology (EMT) especially in analog computing mode is used. Unfortunately, one of the major limitations of EMT is the intrinsic instability of EMT cells, which can cause a significant loss in accuracy. On the other hand, falling back on a digital mode of operation will erode the potential gains. In this work, we propose three optimization techniques that can fully recover the accuracy of EMT-based analog in-memory deep learning models while minimizing their energy consumption: They include the device-enhanced dataset, energy regularization, and low-fluctuation decomposition. Based on the experiment results, we offer two solutions. Developers can either apply the first two optimization techniques or apply all three to the target model. Both solutions can achieve higher accuracy than their state-of-the-art counterparts. The first solution shows at least one order of magnitude improvement in energy efficiency, with the least hardware cost and latency. The second solution further improves energy efficiency by another order of magnitude at the cost of higher latency.

\bibliography{myref}

\begin{thebibliography}{10}
\providecommand{\url}[1]{#1}
\csname url@samestyle\endcsname
\providecommand{\newblock}{\relax}
\providecommand{\bibinfo}[2]{#2}
\providecommand{\BIBentrySTDinterwordspacing}{\spaceskip=0pt\relax}
\providecommand{\BIBentryALTinterwordstretchfactor}{4}
\providecommand{\BIBentryALTinterwordspacing}{\spaceskip=\fontdimen2\font plus
\BIBentryALTinterwordstretchfactor\fontdimen3\font minus
  \fontdimen4\font\relax}
\providecommand{\BIBforeignlanguage}[2]{{%
\expandafter\ifx\csname l@#1\endcsname\relax
\typeout{** WARNING: IEEEtran.bst: No hyphenation pattern has been}%
\typeout{** loaded for the language `#1'. Using the pattern for}%
\typeout{** the default language instead.}%
\else
\language=\csname l@#1\endcsname
\fi
#2}}
\providecommand{\BIBdecl}{\relax}
\BIBdecl

\bibitem{inference2015performance}
G.-B. D.~L. Inference and B.~D. Learning, ``A performance and power analysis,''
  \emph{NVidia Whitepaper, Nov}, 2015.

\bibitem{yang2017designing}
T.-J. Yang, Y.-H. Chen, and V.~Sze, ``Designing energy-efficient convolutional
  neural networks using energy-aware pruning,'' in \emph{Proceedings of the
  IEEE Conference on Computer Vision and Pattern Recognition}, 2017, pp.
  5687--5695.

\bibitem{meena2014}
J.~Meena, S.~Sze, U.~Chand, and T.-Y. Tseng, ``Overview of emerging
  non-volatile memory technologies,'' \emph{Nanoscale Research Letters},
  vol.~9, pp. 1--33, 09 2014.

\bibitem{yao2020fully}
P.~Yao, H.~Wu, B.~Gao, J.~Tang, Q.~Zhang, W.~Zhang, J.~J. Yang, and H.~Qian,
  ``Fully hardware-implemented memristor convolutional neural network,''
  \emph{Nature}, vol. 577, no. 7792, pp. 641--646, 2020.

\bibitem{cai2019fully}
F.~Cai, J.~M. Correll, S.~H. Lee, Y.~Lim, V.~Bothra, Z.~Zhang, M.~P. Flynn, and
  W.~D. Lu, ``A fully integrated reprogrammable memristor--{CMOS} system for
  efficient multiply--accumulate operations,'' \emph{Nature Electronics},
  vol.~2, no.~7, pp. 290--299, 2019.

\bibitem{luo2021nc}
T.~Luo, L.~Yang, H.~Zhang, C.~Qu, X.~Wang, Y.~Cui, W.-F. Wong, and R.~S.~M.
  Goh, ``{NC-Net}: Efficient neuromorphic computing using aggregated sub-nets
  on a crossbar-based architecture with non-volatile memory,'' \emph{IEEE
  Transactions on Computer-Aided Design of Integrated Circuits and Systems},
  2021.

\bibitem{sebastian2020memory}
A.~Sebastian, M.~Le~Gallo, R.~Khaddam-Aljameh, and E.~Eleftheriou, ``Memory
  devices and applications for in-memory computing,'' \emph{Nature
  nanotechnology}, vol.~15, no.~7, pp. 529--544, 2020.

\bibitem{raghavan2013rtn}
N.~Raghavan, R.~Degraeve, L.~Goux, A.~Fantini, D.~Wouters, G.~Groeseneken, and
  M.~Jurczak, ``{RTN} insight to filamentary instability and disturb immunity
  in ultra-low power switching {HfOx} and {AlOx RRAM},'' in \emph{2013
  Symposium on VLSI Technology}.\hskip 1em plus 0.5em minus 0.4em\relax IEEE,
  2013, pp. T164--T165.

\bibitem{luo2018fpga}
T.~Luo, X.~Wang, C.~Qu, M.~K.~F. Lee, W.~T. Tang, W.-F. Wong, and R.~S.~M. Goh,
  ``An {FPGA}-based hardware emulator for neuromorphic chip with {RRAM},''
  \emph{IEEE Transactions on Computer-Aided Design of Integrated Circuits and
  Systems}, vol.~39, no.~2, pp. 438--450, 2018.

\bibitem{du2020exploring}
Y.~Du, L.~Jing, H.~Fang, H.~Chen, Y.~Cai, R.~Wang, J.~Zhang, and Z.~Ji,
  ``Exploring the impact of random telegraph noise-induced accuracy loss on
  resistive {RAM}-based deep neural network,'' \emph{IEEE Transactions on
  Electron Devices}, vol.~67, no.~8, pp. 3335--3340, 2020.

\bibitem{DeepCompression}
S.~Han, H.~Mao, and W.~J. Dally, ``Deep compression: Compressing deep neural
  network with pruning, trained quantization and huffman coding,'' in \emph{4th
  International Conference on Learning Representations, {ICLR} 2016, San Juan,
  Puerto Rico, May 2-4, 2016, Conference Track Proceedings}, 2016.

\bibitem{wang2019haq}
K.~Wang, Z.~Liu, Y.~Lin, J.~Lin, and S.~Han, ``Haq: Hardware-aware automated
  quantization with mixed precision,'' in \emph{Proceedings of the IEEE/CVF
  Conference on Computer Vision and Pattern Recognition}, 2019, pp. 8612--8620.

\bibitem{wang2021evolutionary}
Z.~Wang, T.~Luo, M.~Li, J.~T. Zhou, R.~S.~M. Goh, and L.~Zhen, ``Evolutionary
  multi-objective model compression for deep neural networks,'' \emph{IEEE
  Computational Intelligence Magazine}, vol.~16, no.~3, pp. 10--21, 2021.

\bibitem{emara2014differential}
A.~Emara, M.~Ghoneima, and M.~El-Dessouky, ``Differential {1T2M} memristor
  memory cell for single/multi-bit {RRAM} modules,'' in \emph{2014 6th Computer
  Science and Electronic Engineering Conference (CEEC)}.\hskip 1em plus 0.5em
  minus 0.4em\relax IEEE, 2014, pp. 69--72.

\bibitem{sun2018xnor}
X.~Sun, S.~Yin, X.~Peng, R.~Liu, J.-s. Seo, and S.~Yu, ``{XNOR-RRAM}: A
  scalable and parallel resistive synaptic architecture for binary neural
  networks,'' in \emph{2018 Design, Automation \& Test in Europe Conference \&
  Exhibition (DATE)}.\hskip 1em plus 0.5em minus 0.4em\relax IEEE, 2018, pp.
  1423--1428.

\bibitem{chen201865nm}
W.-H. Chen, K.-X. Li, W.-Y. Lin, K.-H. Hsu, P.-Y. Li, C.-H. Yang, C.-X. Xue,
  E.-Y. Yang, Y.-K. Chen, Y.-S. Chang \emph{et~al.}, ``A 65nm {1Mb} nonvolatile
  computing-in-memory {ReRAM} macro with sub-16ns multiply-and-accumulate for
  binary {DNN AI} edge processors,'' in \emph{2018 IEEE International
  Solid-State Circuits Conference-(ISSCC)}.\hskip 1em plus 0.5em minus
  0.4em\relax IEEE, 2018, pp. 494--496.

\bibitem{tang2017binary}
T.~Tang, L.~Xia, B.~Li, Y.~Wang, and H.~Yang, ``Binary convolutional neural
  network on {RRAM},'' in \emph{2017 22nd Asia and South Pacific Design
  Automation Conference (ASP-DAC)}.\hskip 1em plus 0.5em minus 0.4em\relax
  IEEE, 2017, pp. 782--787.

\bibitem{rastegari2016xnor}
M.~Rastegari, V.~Ordonez, J.~Redmon, and A.~Farhadi, ``Xnor-net: {Imagenet}
  classification using binary convolutional neural networks,'' in
  \emph{European conference on computer vision}.\hskip 1em plus 0.5em minus
  0.4em\relax Springer, 2016, pp. 525--542.

\bibitem{zhu2019configurable}
Z.~Zhu, H.~Sun, Y.~Lin, G.~Dai, L.~Xia, S.~Han, Y.~Wang, and H.~Yang, ``A
  configurable multi-precision cnn computing framework based on single bit
  {RRAM},'' in \emph{2019 56th ACM/IEEE Design Automation Conference
  (DAC)}.\hskip 1em plus 0.5em minus 0.4em\relax IEEE, 2019, pp. 1--6.

\bibitem{peng2019optimizing}
X.~Peng, R.~Liu, and S.~Yu, ``Optimizing weight mapping and data flow for
  convolutional neural networks on {RRAM} based processing-in-memory
  architecture,'' in \emph{2019 IEEE International Symposium on Circuits and
  Systems (ISCAS)}.\hskip 1em plus 0.5em minus 0.4em\relax IEEE, 2019, pp.
  1--5.

\bibitem{he2019noise}
Z.~He, J.~Lin, R.~Ewetz, J.-S. Yuan, and D.~Fan, ``Noise injection adaption:
  End-to-end {ReRAM} crossbar non-ideal effect adaption for neural network
  mapping,'' in \emph{Proceedings of the 56th Annual Design Automation
  Conference 2019}, 2019, pp. 1--6.

\bibitem{chai2018impact}
Z.~Chai, P.~Freitas, W.~Zhang, F.~Hatem, J.~F. Zhang, J.~Marsland,
  B.~Govoreanu, L.~Goux, and G.~S. Kar, ``Impact of {RTN} on pattern
  recognition accuracy of {RRAM}-based synaptic neural network,'' \emph{IEEE
  Electron Device Letters}, vol.~39, no.~11, pp. 1652--1655, 2018.

\bibitem{sorbaro2020optimizing}
M.~Sorbaro, Q.~Liu, M.~Bortone, and S.~Sheik, ``Optimizing the energy
  consumption of spiking neural networks for neuromorphic applications,''
  \emph{Frontiers in neuroscience}, vol.~14, p. 662, 2020.

\bibitem{choi2014random}
S.~Choi, Y.~Yang, and W.~Lu, ``Random telegraph noise and resistance switching
  analysis of oxide based resistive memory,'' \emph{Nanoscale}, vol.~6, no.~1,
  pp. 400--404, 2014.

\bibitem{ielmini2010resistance}
D.~Ielmini, F.~Nardi, and C.~Cagli, ``Resistance-dependent amplitude of random
  telegraph-signal noise in resistive switching memories,'' \emph{Applied
  Physics Letters}, vol.~96, no.~5, p. 053503, 2010.

\bibitem{puglisi2015statistical}
F.~M. Puglisi, P.~Pavan, L.~Larcher, and A.~Padovani, ``Statistical analysis of
  random telegraph noise in {HfO2}-based {RRAM} devices in {LRS},''
  \emph{Solid-State Electronics}, vol. 113, pp. 132--137, 2015.

\bibitem{shim2020two}
W.~Shim, J.-s. Seo, and S.~Yu, ``Two-step write--verify scheme and impact of
  the read noise in multilevel {RRAM}-based inference engine,''
  \emph{Semiconductor Science and Technology}, vol.~35, no.~11, p. 115026,
  2020.

\bibitem{joshi2020accurate}
V.~Joshi, M.~Le~Gallo, S.~Haefeli, I.~Boybat, S.~R. Nandakumar, C.~Piveteau,
  M.~Dazzi, B.~Rajendran, A.~Sebastian, and E.~Eleftheriou, ``Accurate deep
  neural network inference using computational phase-change memory,''
  \emph{Nature communications}, vol.~11, no.~1, pp. 1--13, 2020.

\bibitem{zhang2020reliable}
G.~L. Zhang, B.~Li, Y.~Zhu, S.~Zhang, T.~Wang, Y.~Shi, T.-Y. Ho, H.~Li, and
  U.~Schlichtmann, ``Reliable and robust {RRAM}-based neuromorphic computing,''
  in \emph{Proceedings of the 2020 on Great Lakes Symposium on VLSI}, 2020, pp.
  33--38.

\bibitem{joksas2020committee}
D.~Joksas, P.~Freitas, Z.~Chai, W.~Ng, M.~Buckwell, C.~Li, W.~Zhang, Q.~Xia,
  A.~Kenyon, and A.~Mehonic, ``Committee machines—a universal method to deal
  with non-idealities in memristor-based neural networks,'' \emph{Nature
  communications}, vol.~11, no.~1, pp. 1--10, 2020.

\bibitem{wan2020voltage}
W.~Wan, R.~Kubendran, B.~Gao, S.~Josbi, P.~Raina, H.~Wu, G.~Cauwenberghs, and
  H.-S.~P. Wong, ``A voltage-mode sensing scheme with differential-row weight
  mapping for energy-efficient {RRAM}-based in-memory computing,'' in
  \emph{2020 IEEE Symposium on VLSI Technology}.\hskip 1em plus 0.5em minus
  0.4em\relax IEEE, 2020, pp. 1--2.

\bibitem{pedretti2021memory}
G.~Pedretti and D.~Ielmini, ``In-memory computing with resistive memory
  circuits: Status and outlook,'' \emph{Electronics}, vol.~10, no.~9, p. 1063,
  2021.

\bibitem{wang2020ncpower}
Z.~Wang, H.~Zhang, T.~Luo, W.-F. Wong, A.~T. Do, P.~Vishnu, W.~Zhang, and
  R.~S.~M. Goh, ``{NCPower}: Power modelling for {NVM}-based neuromorphic
  chip,'' in \emph{International Conference on Neuromorphic Systems 2020},
  2020, pp. 1--7.

\bibitem{gong2018signal}
N.~Gong, T.~Id{\'e}, S.~Kim, I.~Boybat, A.~Sebastian, V.~Narayanan, and
  T.~Ando, ``Signal and noise extraction from analog memory elements for
  neuromorphic computing,'' \emph{Nature communications}, vol.~9, no.~1, pp.
  1--8, 2018.

\bibitem{bottou2012stochastic}
L.~Bottou, ``Stochastic gradient descent tricks,'' in \emph{Neural networks:
  Tricks of the trade}.\hskip 1em plus 0.5em minus 0.4em\relax Springer, 2012,
  pp. 421--436.

\bibitem{zhang2018improved}
Z.~Zhang, ``Improved {ADAM} optimizer for deep neural networks,'' in \emph{2018
  IEEE/ACM 26th International Symposium on Quality of Service (IWQoS)}.\hskip
  1em plus 0.5em minus 0.4em\relax IEEE, 2018, pp. 1--2.

\bibitem{lee2019system}
M.~K.~F. Lee, Y.~Cui, T.~Somu, T.~Luo, J.~Zhou, W.~T. Tang, W.-F. Wong, and
  R.~S.~M. Goh, ``A system-level simulator for {RRAM}-based neuromorphic
  computing chips,'' \emph{ACM Transactions on Architecture and Code
  Optimization (TACO)}, vol.~15, no.~4, pp. 1--24, 2019.

\bibitem{pytorchzoo}
``Pytorch model library,'' pytorch.org/vision/stable/models.html, accessed:
  2021-10-23.

\bibitem{raghavan2013microscopic}
N.~Raghavan, R.~Degraeve, A.~Fantini, L.~Goux, S.~Strangio, B.~Govoreanu,
  D.~Wouters, G.~Groeseneken, and M.~Jurczak, ``Microscopic origin of random
  telegraph noise fluctuations in aggressively scaled {RRAM} and its impact on
  read disturb variability,'' in \emph{2013 IEEE International Reliability
  Physics Symposium (IRPS)}.\hskip 1em plus 0.5em minus 0.4em\relax IEEE, 2013,
  pp. 5E--3.

\end{thebibliography}
\bibliographystyle{IEEEtran}

\end{document}